\documentclass{article}

\PassOptionsToPackage{numbers, sort&compress}{natbib}

\usepackage[preprint]{neurips_2020}

\usepackage[utf8]{inputenc} 
\usepackage[T1]{fontenc}    
\usepackage{hyperref}       
\usepackage{url}            
\usepackage{booktabs}       
\usepackage{amsfonts}       
\usepackage{nicefrac}       
\usepackage{microtype}      
\usepackage{graphicx}
\usepackage[caption=false,font=footnotesize]{subfig}
\usepackage{comment}
\usepackage{amsmath,amssymb} 
\usepackage{color}
\usepackage{multirow}
\usepackage{todonotes}
\usepackage[capitalize]{cleveref}
\usepackage{tabularx}
\usepackage[detect-all=true,separate-uncertainty=true]{siunitx}
\usepackage{wrapfig}
\usepackage{array}

\usepackage{multibib}
\newcites{A}{References}

\newcommand{\erw}{\mathbb{E}}
\newcommand{\var}{\mathbb{V}}

\newcommand{\calD}{\mathcal{D}}

\newcommand{\calN}{\mathcal{N}}

\newcommand{\bbR}{\mathbb{R}}

\newcommand{\eps}{\varepsilon}

\newcommand*\diff{\mathop{}\!\mathrm{d}}

\DeclareMathOperator{\KL}{KL}
\DeclareMathOperator{\ELBO}{ELBO}
\DeclareMathOperator{\ReLU}{ReLU}
\DeclareMathOperator{\softmax}{softmax}

\newcolumntype{Y}{>{\centering\arraybackslash}X}

\makeatletter
\ifcase \@ptsize \relax
  \newcommand{\miniscule}{\@setfontsize\miniscule{4}{5}}
\or
  \newcommand{\miniscule}{\@setfontsize\miniscule{5}{6}}
\or
  \newcommand{\miniscule}{\@setfontsize\miniscule{5}{6}}
\fi
\makeatother

\title{Sampling-free Variational Inference for Neural Networks with Multiplicative Activation Noise}

\author{%
  Jannik Schmitt \hspace{3em} Stefan Roth \\
  Dept. of Computer Science, TU Darmstadt\\
  \texttt{\{jannik.schmitt, stefan.roth\}@visinf.tu-darmstadt.de}
}

\begin{document}

\maketitle

\begin{abstract}
  To adopt neural networks in safety critical domains, knowing whether we can trust their predictions is crucial.
  Bayesian neural networks (BNNs) provide uncertainty estimates by averaging predictions with respect to the posterior weight distribution.
  Variational inference methods for BNNs approximate the intractable weight posterior with a tractable distribution, yet mostly rely on sampling from the variational distribution during training and inference.
  Recent sampling-free approaches offer an alternative, but incur a significant parameter overhead.
  We here propose a more efficient parameterization of the posterior approximation for sampling-free variational inference that relies on the distribution induced by multiplicative Gaussian activation noise.
  This allows us to combine parameter efficiency with the benefits of sampling-free variational inference.
  Our approach yields competitive results for standard regression problems and scales well to large-scale image classification tasks including ImageNet. 
\end{abstract}

\section{Introduction}

When applying deep networks to safety critical problems, uncertainty estimates for their prediction are paramount.
Bayesian inference is a theoretically well-founded framework for estimating the model-inherent uncertainty by computing the posterior distribution of the parameters.
While sampling from the posterior of a Bayesian neural network is possible with different Markov chain Monte Carlo (MCMC) methods, often an explicit approximation of the posterior can be beneficial, for example for continual learning \cite{nguyen2018variational}.
Variational inference (VI) can be used to approximate the posterior with a simpler distribution.
Since the variational objective as well as the predictive distribution cannot be calculated analytically, they are often approximated through Monte Carlo integration using samples from the approximate posterior \cite{graves2011practical, blundell2015weight}.
During training, this introduces additional gradient variance which can be a problem when training large BNNs.
Further, multiple forward-passes are required to compute the predictive distribution, which makes deployment in time-critical systems difficult.

Recently, sampling-free variational inference methods \cite{roth2016variational, kandemir2018sampling, wu2019deterministic} have been proposed, with similar predictive performance to sampling-based VI methods on small-scale tasks.
They may also be able to remedy the gradient variance problem of sampling-based VI for larger-scale tasks.
Still, sampling-free methods incur a significant parameter overhead.
To address this, we propose a sampling-free variational inference scheme for BNNs -- termed MNVI -- where the approximate posterior can be induced by \emph{multiplicative Gaussian activation noise}.
This helps us \emph{decrease the number of parameters of the Bayesian network to almost half}, while still being able to analytically compute the Kullback-Leibler (KL) divergence with regard to an isotropic Gaussian prior.
Further, assuming multiplicative activation noise allows to reduce the computational cost of variance propagation in Bayesian networks compared to a Gaussian mean-field approximate posterior.
We then discuss how our MNVI method can be applied to modern network architectures with batch normalization and max-pooling layers.
Finally, we describe how regularization by the KL-divergence term differs for networks with the induced variational posterior from networks with a mean-field variational posterior.

In experiments on standard regression tasks \cite{Dua:2019}, our proposed sampling-free variational inference method achieves competitive results while being more lightweight than other sampling-free methods.
We further apply our method to large-scale image classification problems using modern convolutional network architectures including ResNet \cite{he2016deep}, obtaining well-calibrated uncertainty estimates while also improving the prediction accuracy compared to standard deterministic networks.

\begin{figure}[t]
\centering
\subfloat[Standard sampling-free variational inference\label{fig:teaser-std}]{%
  \centering
  \hspace{0.06\linewidth}
  \includegraphics[width=.35\linewidth,trim=0 10 0 10]{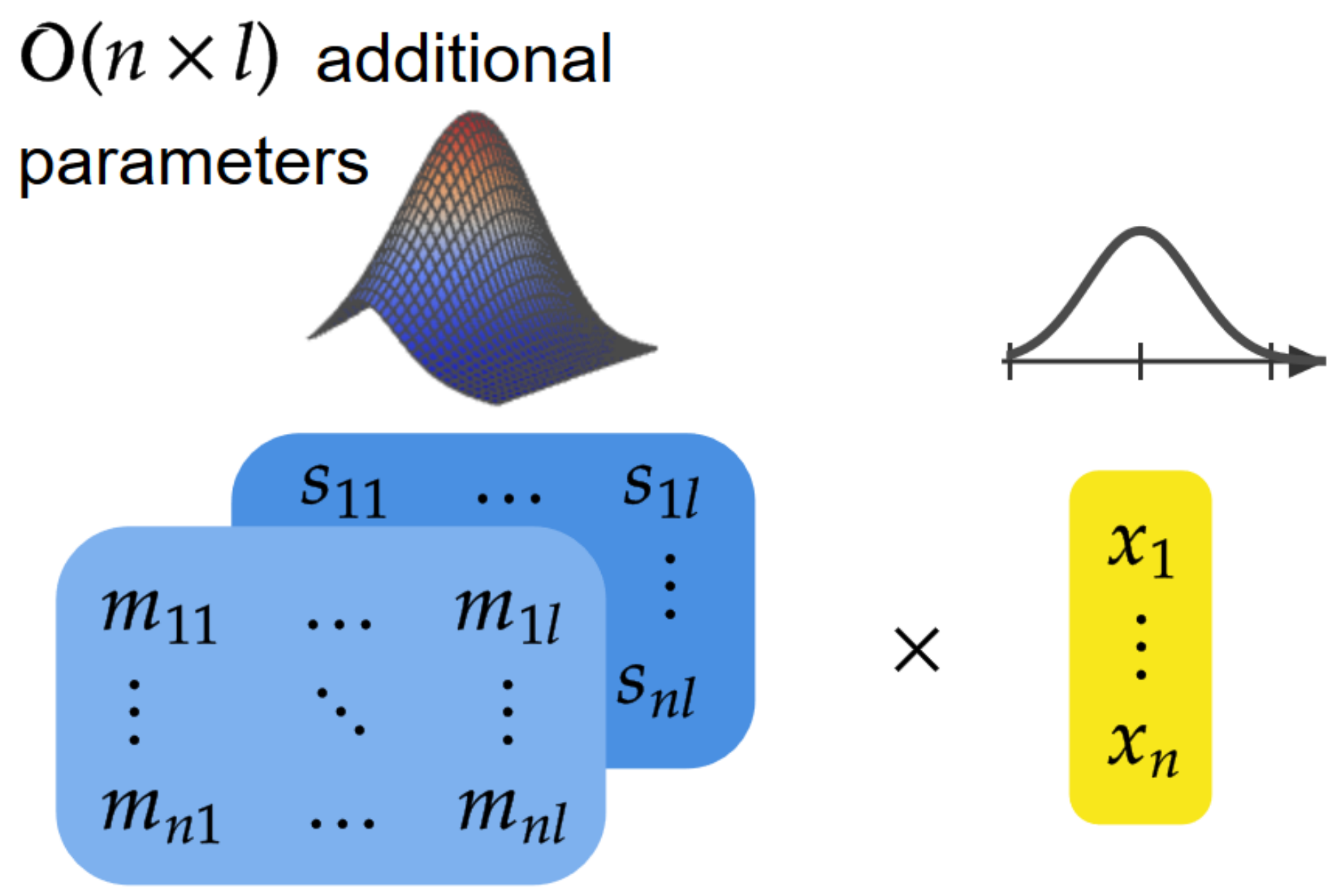}
  \hspace{0.06\linewidth}
}
\subfloat[Our MNVI approach\label{fig:teaser-ours}]{%
  \centering
  \hspace{0.06\linewidth}
  \includegraphics[width=.35\linewidth,trim=0 10 0 10]{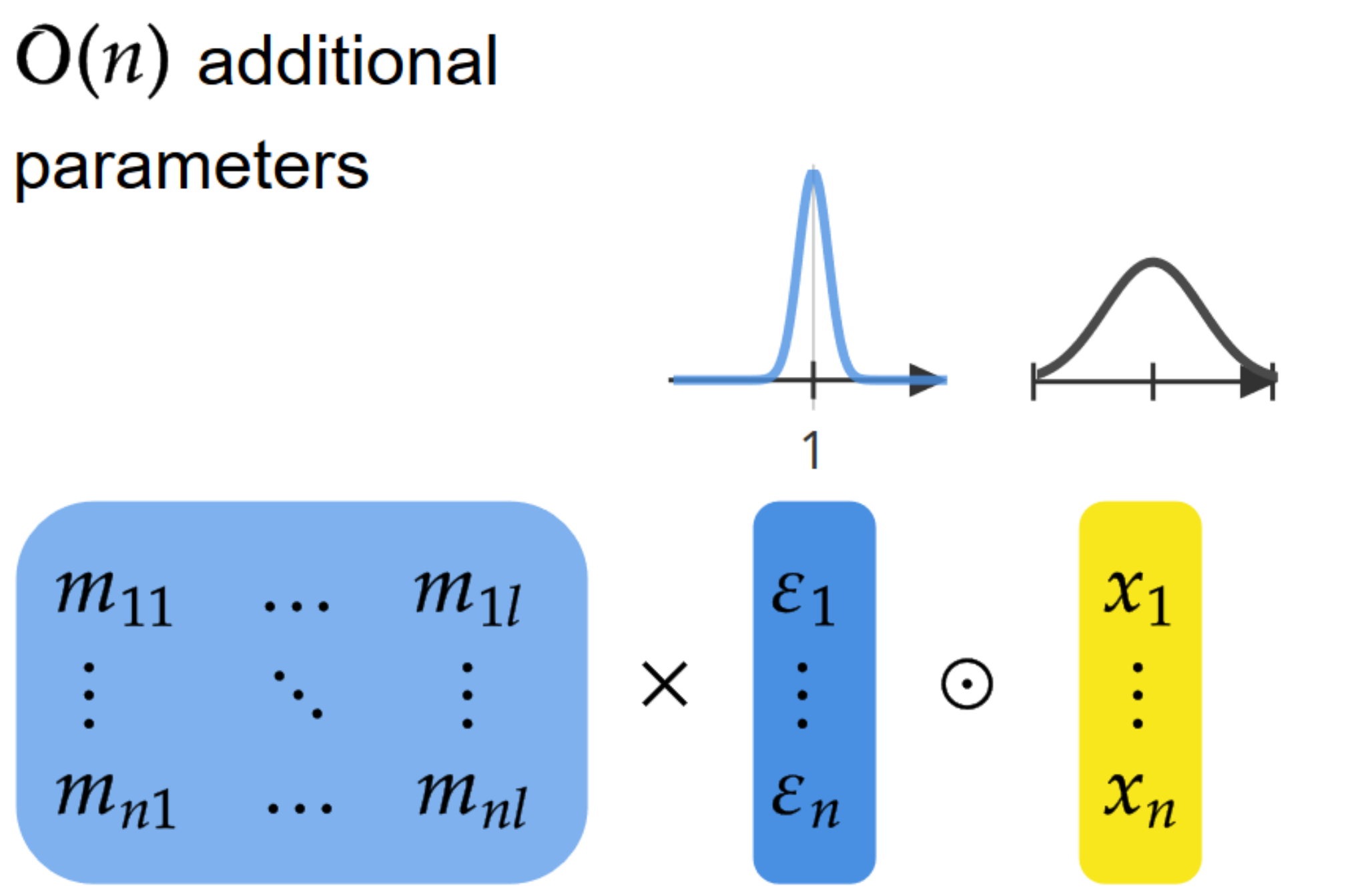}
  \hspace{0.06\linewidth}
}
\vspace{-1mm}
\caption{Both standard sampling-free mean-field VI \emph{\protect\subref{fig:teaser-std}} and our MNVI approach \emph{\protect\subref{fig:teaser-ours}} assume that the input $x$ to a linear layer with weight matrix $M$ as well as the weight uncertainty of the posterior can be approximated by Gaussian distributions. 
We additionally assume that the variational posterior is induced by multiplicative Gaussian activation noise $\eps$ with mean 1.
This lets us represent the weight uncertainty \emph{compactly} as an $n$-dimensional vector, storing the variance of the activation noise. 
In contrast, a mean-field variational posterior needs an $n\times l$  matrix $S$ to store the weight variances.}
\label{fig:teaser}
\vspace{-2mm}
\end{figure}

We make the following contributions: \emph{(i)} We propose to  reduce the number of parameters and computations for sampling-free variational inference by using the distribution induced by multiplicative Gaussian activation noise in neural networks as a variational posterior (cf.~\cref{fig:teaser});
\emph{(ii)}~we show how our MNVI method can be applied to common network architectures that are used in practice;
\emph{(iii)}~we demonstrate experimentally that our method retains the accuracy of sampling-free VI with a mean-field Gaussian variational posterior while performing better with regard to uncertainty specific evaluation metrics despite fewer parameters; and
\emph{(iv)}~we successfully apply it to various large-scale image classification problems such as classification on ImageNet. 

\section{Related Work}

\textbf{Bayesian neural networks} (BNNs) allow for a principled quantification of model uncertainty in neural networks. Early approaches for learning the posterior distribution of BNNs rely on MCMC methods \cite{neal1995bayesian} or a Laplace approximation \cite{mackay1992practical} and are difficult to scale to modern neural networks.
More modern approaches use a variety of Bayesian inference methods, such as assumed density filtering for probabilistic backpropagation \cite{hernandez2015probabilistic,ghosh2016assumed}, factorized Laplace approximation \cite{ritter2018scalable}, stochastic gradient MCMC methods like SGLD \cite{welling2011bayesian}, IASG \cite{mandt2017stochastic}, or more recent methods \cite{heek2019bayesian, zhang2019cyclical}, as well as variational inference.

\textbf{Variational inference} for neural networks was first proposed by \citet{hinton1993keeping}, motivated from an information-theoretic perspective.
Further stochastic descent-based VI algorithms were developed by \citet{graves2011practical} and \citet{blundell2015weight}.
Latter utilizes the local re-parametrization trick \cite{Kingma14auto} and can be extended to variational versions \cite{khan2018fast} of adaptive gradient-based optimization with preconditioning depending on the gradient noise, such as Adam \cite{kingma2015adam} and RMSProp \cite{hinton2012neural}. \citet{swiatkowski2020k} recently proposed an efficient low-rank parameterization of the mean-field posterior.

\textbf{Multiplicative activation noise}, such as Bernoulli or Gaussian noise, is well-known from the regularization methods Dropout and Gaussian Dropout \cite{srivastava2014dropout}. \citet{gal2016dropout} and \citet{kingma2015variational} showed that Dropout and Gaussian Dropout, respectively, can be interpreted as approximate VI.
Since for both methods the multiplicative noise is sampled once per activation and not for each individual weight, the induced posterior is low-dimensional and the evidence lower bound (ELBO) is not well-defined \cite{hron2018variational}.
Rank-1 BNNs \cite{dusenberry2020efficient} use multiplicative noise with hierarchical priors to perform sampling-based VI.
More complex distributions for the multiplicative noise modeled by normalizing flows were studied by \citet{louizos2017multiplicative}.
Multiplicative noise has further been considered for reducing sampling variance by generating pseudo-independent weight samples \cite{wen2018flipout} and to address the increase of parameters for ensembles \cite{wen2019batchensemble}.

\textbf{Variance propagation} for sampling-free VI has been proposed by \citet{roth2016variational}, \citet{kandemir2018sampling}, and \citet{wu2019deterministic}.
\citet{jankowiak2018closed} computes closed form objectives in the case of a single hidden layer.
Variance propagation has also been used in various uncertainty estimation algorithms for estimating epistemic uncertainty by Bayesian principles, such as Probabilistic Back Propagation \cite{hernandez2015probabilistic}, Stochastic Expectation Propagation \cite{li2015stochastic}, and Neural Parameter Networks \cite{wang2016natural}, or aleatoric uncertainty by propagating an input noise estimate through the network \cite{jin2015robust,gast2018lightweight}.

Most related to our approach, multiplicative activation noise and variance propagation have been jointly used by \citet{wang2013fast} for training networks with dropout noise without sampling, as well as by \citet{postels2019sampling} for sampling-free uncertainty estimation for networks that are trained by optimizing the approximate ELBO of \citep{gal2016dropout}.
However, \citep{wang2013fast} estimates a different objective than the ELBO in variational inference and does not estimate predictive uncertainty, while \citep{postels2019sampling} uses Monte Carlo Dropout during training.
Further, both methods assume Bernoulli noise on the network's activation that is not adapted during training.
Our approach not only retains the low parameter overhead of dropout-based methods, but offers a well-defined ELBO.
To the best of our knowledge, we are the first to perform sampling-free VI for networks with adaptive Gaussian activation noise.

\section{MNVI -- Variational Inference in Neural Networks with Multiplicative Gaussian Activation Noise}

\subsection{Variational inference and the evidence lower bound}

Given a likelihood function $p(y | x, w)$ parameterized by a neural network with weights $w$, data $\calD=(X, Y)$, and a prior on the network's weights $p(w)$, we are interested in the posterior distribution of the network weights
\begin{equation}
  p(w | X,Y) = \frac{1}{p(Y|X)}p(Y|X,w)p(w).
\end{equation}
For neural networks, however, computing $p(Y|X)$ and the exact posterior $p(w|X,Y)$ is intractable.
A popular method to approximate the posterior is variational inference.
Let $q(w | \theta)$ be a family of tractable distributions parameterized by $\theta$.
The objective of variational inference \cite{wainwright2008graphical} is to choose a distribution $q(w | \theta^*)$ such that $\theta^*$ minimizes $\KL\big[q(w|\theta)\big\|p(w|X,Y)\big]$,
where $\KL[\cdot\|\cdot]$ denotes the Kullback-Leibler divergence.
Minimizing the divergence is equivalent to maximizing the evidence lower bound
\begin{equation}
  \ELBO(\theta) = \erw_{q(w|\theta)}\big[\log p(Y|X,w)\big] - \KL\big[q(w|\theta)\big\|p(w)\big],
\end{equation}
where $\erw_{q(w|\theta)}[\log p(Y|X,w)]$ is the expected log-likelihood with respect to the variational distribution and $\KL[q(w|\theta)||p(w)]$ is the divergence of the variational distribution from the prior.
Finding a distribution that maximizes the ELBO can, therefore, be understood as a trade-off between fitting to a data-dependent term while not diverging too far from the prior belief about the weight distribution.

\subsection{The implicit posterior of BNNs with adaptive Gaussian activation noise}

For variational inference in Bayesian neural networks, the expected log-likelihood cannot be calculated exactly and has to be approximated, commonly through Monte Carlo integration with regards to samples from the variational distribution \cite{blundell2015weight, graves2011practical}:
\begin{equation}
  \erw_{q(w|\theta)}\big[\log p(Y|X,w)\big] \approx \frac{1}{S}\sum_{s=1}^S \log p(Y|X,w^s), \quad w^s \sim q(w|\theta).
\end{equation}
\citet{wu2019deterministic} showed that in case the output distribution $q(h(X))$ of a function $h$ modeled by a neural network is known, the expected value can instead be computed by integration with respect to the output distribution:
\begin{equation}
  \erw_{q(w|\theta)}\big[\log p(Y|X,w)\big] = \erw_{q(h(X))}\big[\log p(Y|h(X))\big].
\end{equation}
An alternative way to approximate the expected log-likelihood is, therefore, to approximate the output distribution using variance propagation.

The variational family used to approximate the weight posterior of a BNN is often chosen to be Gaussian. 
However, even a naive Gaussian mean-field approximation with a diagonal covariance matrix doubles the number of parameters compared to a deterministic network, which is problematic for large-scale networks that are relevant in various application domains.

\textbf{Assumption 1.}
To reduce the parameter overhead, \emph{we here propose to use the induced Gaussian variational distribution of a network with multiplicative Gaussian activation noise} \cite{kingma2015variational}.
Given the output $x_j$ of an activation function with independent multiplicative $\calN(1,\alpha_j)$-distributed noise $\eps_{ij}$ and a weight $m_{ij}$, the product $m_{ij}\eps_{ij}x_j$ is distributed equally to the product of $x_j$ and a stochastic $\calN(m_{ij}, \alpha_j m_{ij}^2)$-distributed weight $w_{ij}$.
Therefore, the implicit variational distribution of networks with independent multiplicative Gaussian activation noise is given by
\begin{equation}
  q(w|m, \alpha) = \prod_{i,j} \calN(w_{ij}|m_{ij}, \alpha_j m_{ij}^2).
\end{equation}
This distribution is parameterized by the weight means $m_{ij}$ and activation noise $a_j$, hence for a linear layer with $M_1$ input units and $M_2$ output units, it introduces only $M_1$ new parameters in addition to the $M_1 \times M_2$ parameters of a deterministic model.
This way, the quadratic scaling of the number of additional parameters with respect to the network width for a traditional mean-field variational posterior with a diagonal covariance matrix approximation \cite{graves2011practical, wu2019deterministic} can be \emph{reduced to linear scaling} in our case, as visualized in \cref{fig:teaser}.
Thus our assumption significantly reduces the parameter overhead.
As we will see, this not only benefits practicality, but in fact even improves uncertainty estimates.

Further for convolutional layers, we assume that the variance parameter of the multiplicative noise is shared within a channel. For a convolutional layer with $C_1$ input channels, $C_2$ output channels, and filter width $W$, only $C_1$ new parameters are introduced in addition to the $C_1 \times C_2 \times W^2$ parameters of the deterministic layer.
While the number of additional parameters for a mean-field variational posterior scales quadratic with respect to the number of channels $C_1 \times C_2$ and filter width $W$, the number of additional parameters for the distribution induced by our multiplicative activation noise assumption scales only \emph{linearly} in $C_1$, again significantly reducing the parameter overhead.

\subsection{Variance propagation}
Assuming we can compute the distribution of a network's output $q(h(x))$, which is induced by parameter uncertainties, as well as $\erw_{q(h(X))}[\log p(Y|h(X))]$ and $\KL[q(w|\theta)||p(w)]$ for the chosen variational family and prior distribution, we can perform sampling-free variational inference \cite{wu2019deterministic}.
To approximate the output distribution $q(h(x))$, variance propagation can be utilized.
Variance propagation can be made tractable and efficient by two assumptions:

\textbf{Assumption 2.} \textit{The output of a linear or convolutional layer is approximately Gaussian.}

Since a multivariate Gaussian is characterized by its mean vector and covariance matrix, the approximation of the output distribution of a linear layer can be  calculated by just computing the first two moments.
For sufficiently wide networks this assumption can be justified by the central limit theorem.

\textbf{Assumption 3.} \textit{The correlation of the outputs of linear and convolutional layers can be neglected.}

This assumption is crucial for reducing the computational cost, since it allows our MNVI approach to propagate only the variance vectors instead of full covariance matrices, thus avoiding quadratic scaling with regards to layer width.
While multiplication with independent noise reduces the correlation of the outputs of linear layers, the correlation will generally be non-zero.
However, empirically \citet{wu2019deterministic} verified that the predictive performance of a BNN is mostly retained under this assumption.

\subsubsection{Reducing the computation}
For two independent random variables $\omega$ and $X$ with finite second moment, mean and variance of their product can be calculated by
\begin{align}
    \erw[\omega X] &= \erw[\omega] \erw[X], &
    \var[\omega X] &= \var[\omega] \var[X] + \var[\omega] \erw[X]^2 + \erw[\omega]^2 \var[X].\\
\intertext{Given a linear layer with independent $\calN(m_{ij}, s_{ij})$-distributed weights and independent inputs with means $x_{j}$ and variances $v_{j}$, we can calculate the mean and variance of the outputs $Z$ as}
    \erw[Z_i] &= \sum_j m_{ij} x_j, &
    \var[Z_i] &= \sum_j s_{ij} v_j  + s_{ij} x_j^2 + m_{ij}^2 v_j,\\
\intertext{which can be vectorized as}
    \erw[Z] &= Mx, &
    \var[Z] &= S(v + x \circ x) + (M \circ M) v,
\end{align}
where $\circ$ is the Hadamard-product.
Hence, in addition to one matrix-vector multiplication for the mean propagation, two more matrix-vector multiplications are needed for variance propagation through a linear layer, tripling the computational cost compared to its deterministic counterpart.

Importantly, in the case of our proposed posterior weight distribution for networks with multiplicative activation noise, however, the weight variance $s_{ij}=\alpha_j m_{ij}^2$ can be linked to the weight mean, so that the output variance for our MNVI approach can be computed as
\begin{equation}
  \var[Z] = (M \circ M) (\alpha \circ (v + x \circ x))  + (M \circ M) v = (M \circ M)((1 + \alpha)\circ v + \alpha \circ x \circ x).
\end{equation}
Therefore, variance propagation through linear layers for multiplicative noise networks only requires computing one matrix-vector product and the cost for computing the variance can be reduced by half.

Similarly, for convolutional layers with $\calN(m_{ij}, s_{ij})$-distributed weights in addition to one convolutional operation for mean propagation, generally two additional convolutional operations are needed for variance propagation.
By linking the weight variance to the weight mean with multiplicative activation noise, i.e. $w_{ij} \sim \calN(m_{ij}, \alpha_j m_{ij}^2)$, the cost of variance propagation through a convolutional layer can also be reduced to only one additional convolutional operation.

\subsubsection{Variance propagation through activation layers}

For non-linear activation functions $f$, we assume that the input $Z$ can be approximated Gaussian with mean $\mu$ and variance $\sigma^2$.
We thus calculate the mean and variance of the output distribution as
\begin{equation}
    \erw\big[f(Z)\big] = \int_\bbR f(z) \phi\left(\frac{z-\mu}{\sigma} \right) \diff z, \quad
    \var\big[f(Z)\big] = \int_\bbR f(z)^2 \phi\left(\frac{z-\mu}{\sigma} \right) \diff z - \erw\big[f(Z)\big]^2,
\end{equation}
where $\phi(z)$ is the standard Gaussian probability density function.

As an example, for the rectifier function, mean and variance can be computed analytically \cite{frey1999variational} as
\begin{equation}
      \erw[\ReLU(Z)] = \mu \Phi\left(\frac{\mu}{\sigma}\right) + \sigma \phi\left(\frac{\mu}{\sigma}\right),\
      \var[\ReLU(Z)] = (\mu^2 + \sigma^2) \Phi\left(\frac{\mu}{\sigma}\right) + \mu \sigma \phi\left(\frac{\mu}{\sigma}\right) - \erw[\ReLU(z)]^2.
\end{equation}
If these integrals cannot be computed analytically for some other activation function, they can instead be approximated by first-order Taylor expansion, yielding
\begin{equation}
    \erw\big[f(Z)\big] \approx f(\erw[Z]), \quad \var\big[f(Z)\big] \approx \left. \frac{\diff f(z)}{\diff z}\right\rvert_{\erw[Z]}^2 \cdot \var[Z].
\end{equation}

\subsubsection{Batch normalization and max-pooling layers}

To apply our approach to modern neural networks, variance estimates have to be propagated through batch-normalization \cite{Ioffe15batch} and pooling layers.
Similar to \citet{osawa2019practical}, we aim to retain the stabilizing effects of batch normalization during training and, therefore, do not model a distribution for the batch-normalization parameters and do not change the update rule for the input mean $x$.
For propagating the variance $v$, we view the batch-normalization layer as a linear layer, obtaining 
\begin{equation}
    x' = \alpha \cdot \frac{x-\text{mean(x)}}{\sqrt{\text{var}(x)+\eps}} + \beta, \quad
    v' = \frac{\alpha^2 v}{\text{var}(x)+\eps},
\end{equation}
where $\text{mean}(x)$ and $\text{var}(x)$ are the batch mean and variance, $\alpha, \beta$ are the affine batch-normalization parameters, and $\eps$ is a stabilization constant.

The output mean and variance of the max-pooling layer under the assumption of independent Gaussian inputs cannot be calculated analytically for more then two variables.
Therefore, we use a simple approximation similar to \citet{kandemir2018sampling} that preserves the sparse activations of the deterministic max-pooling layer.
Given an input window of variables with means $(x_{ij})$ and variances $(v_{ij})$, the max-pooling layer outputs $x_{ij^*} = \max_{ij} x_{ij}$ and the variance of the respective entry $v_{ij^*}$.

\subsection{The NLLH term}
Regression problems are often formulated as $L_2$-loss minimization, which is equivalent to maximizing the log-likelihood under the assumption of additive homoscedastic Gaussian noise.
We obtain a more general loss function, if we allow for heteroscedastic noise, which can be modeled by a network predicting an input-dependent estimate of the log-variance $c(x)$ in addition to the predicted mean $\mu(x)$ of the Gaussian.
The log-likelihood of a data point $(x,y)$ under these assumptions is
\begin{equation}
    \log p\big(y|\mu(x), c(x)\big) =  - \frac{1}{2} \left( \log 2 \pi + c(x) + e^{-c(x)} (\mu(x)-y)^2 \right).
\end{equation}
Given a Gaussian approximation for the output distribution of the network, the expected log-likelihood can now be computed analytically.
To represent heteroscedastic aleatoric uncertainty, we assume that the network predicts the mean $\mu(x)$ and the log-variance $c(x)$ of the Gaussian conditional distribution and additionally outputs the variances $v_\mu(x)$ and  $v_c(x)$ of those estimates due to the weight uncertainties represented by the multiplicative Gaussian activation noise.
The expected log-likelihood can then be computed by integrating the log-likelihood with respect to the output distribution $q(h(x))$ of the network and is given by \cite{wu2019deterministic}
\begin{equation}
  \erw_{q(w|\theta)}\big[\log p(y|x,w)\big] = - \frac{1}{2} \left( \log 2 \pi + c(x) + e^{-c(x) + v_c(x)}\big(v_\mu(x) + (\mu(x)-y)^2\big) \right).
\end{equation}

For classification, however, the expected log-likelihood cannot be calculated analytically. Instead, it can be approximated by Monte Carlo integration by sampling logits from the Gaussian approximation of the network's output distribution $\tilde q(h(x)) = \calN(\mu(x), v_\mu(x))$:
\begin{equation}
  \erw_{q(w|\theta)}\big[\log p(y| x, w)\big] \approx \frac{1}{S} \sum_{s=1}^S \log p\big(y|\softmax(h^s(x))\big), \quad h^s(x) \sim \tilde q(h(x)).
\end{equation}
Note that computing this Monte Carlo estimate only requires sampling from the output distribution, which has negligible computational costs compared to sampling predictions by drawing from the weight distribution and computing multiple forward-passes.
Thus our MNVI variational inference scheme is efficient despite this sample approximation.
This basic approach can also be used to compute the expected log-likelihood for a general probability density function.

\subsection{The KL term}
\begin{wrapfigure}[13]{r}{0.5\textwidth}
\vspace{-2mm}
\centering
\includegraphics[width=.4\linewidth]{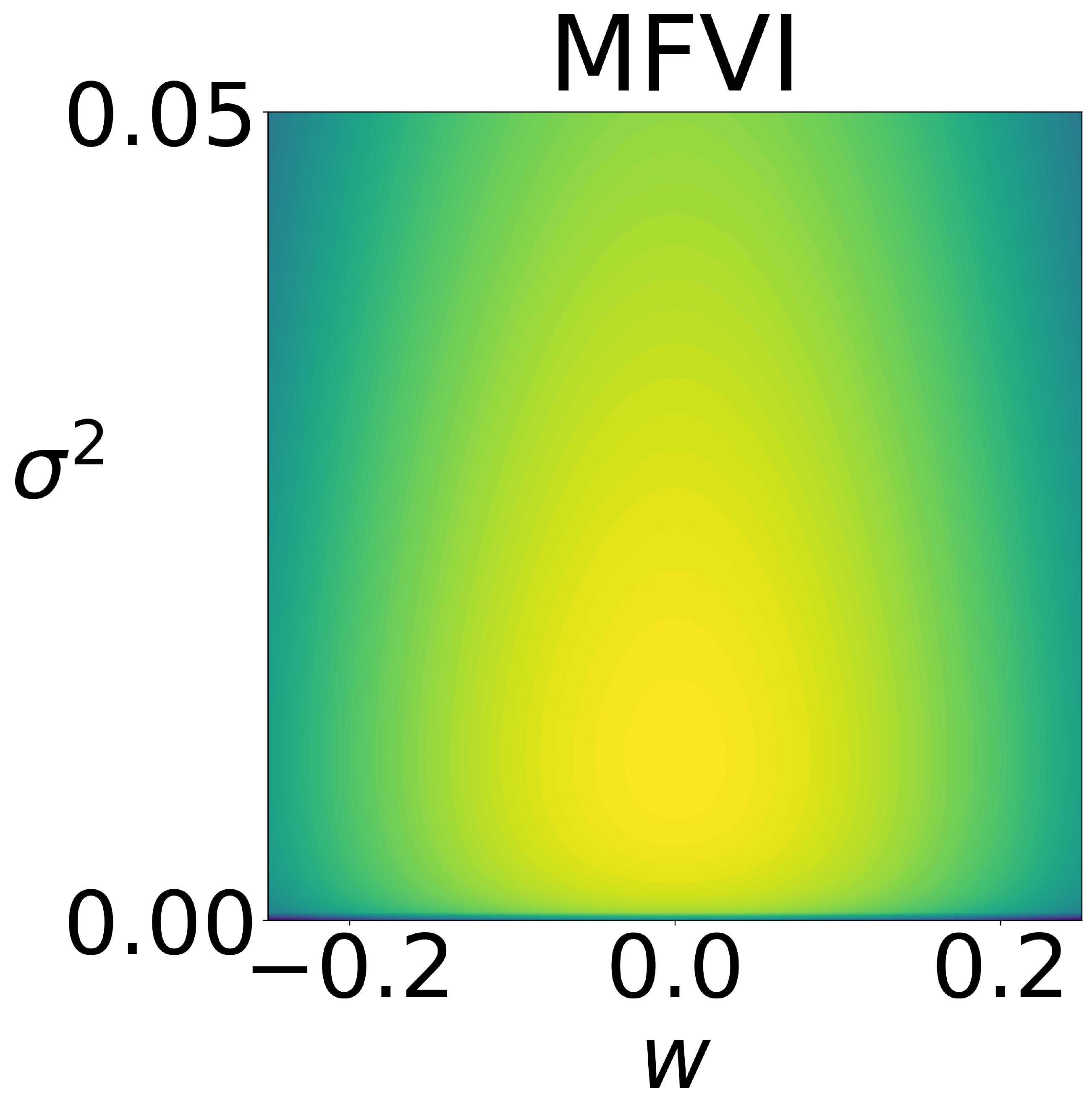}
\hspace{5mm}
\includegraphics[width=.4\linewidth]{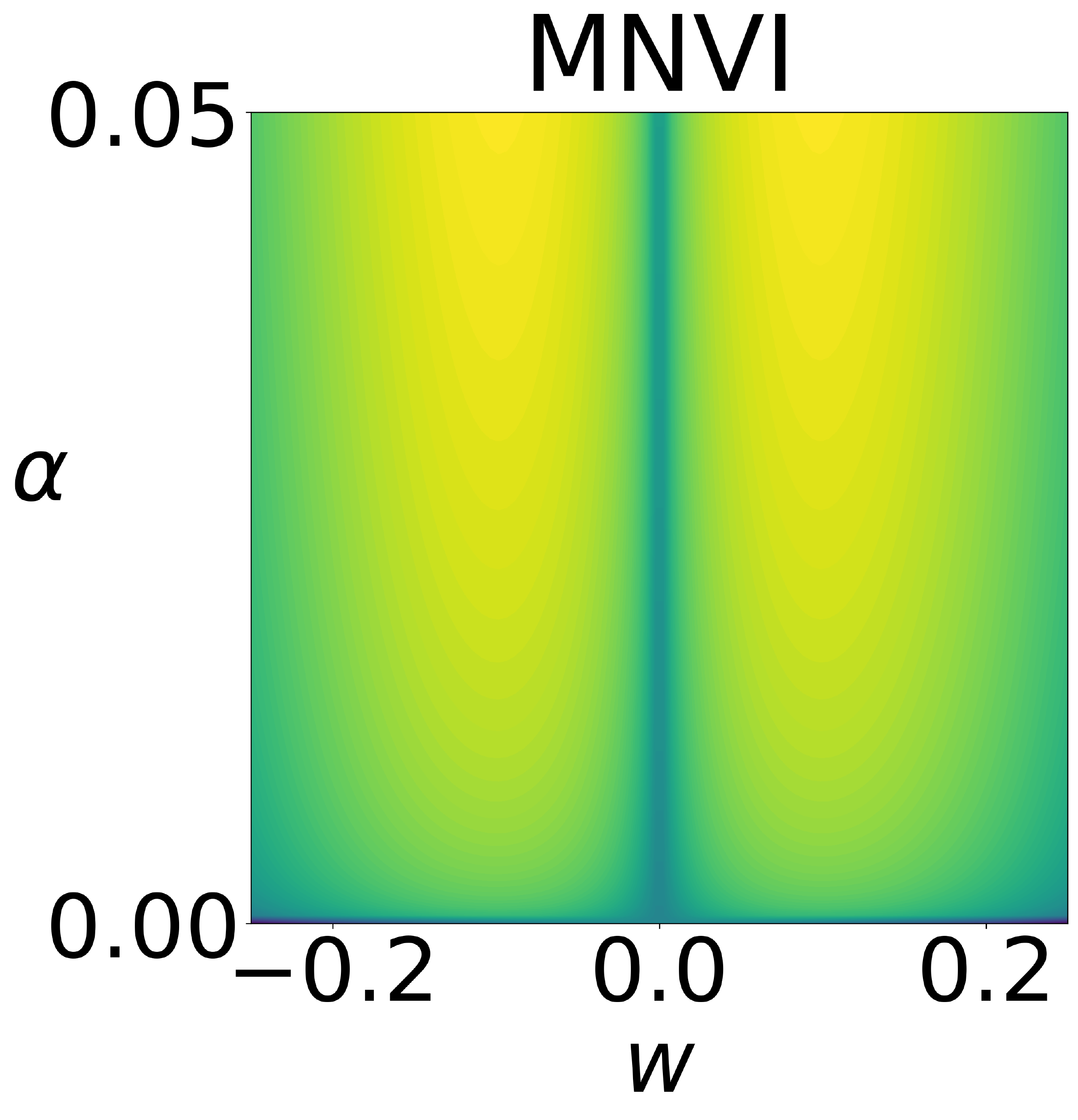}
\vspace{-1.5mm}
\caption{KL divergence from a Gaussian prior for a one-dimensional mean-field Gaussian variational posterior \emph{(left)} and a variational posterior induced by our multiplicative activation noise \emph{(right)}. Brighter areas correspond to a lower divergence.}
\label{fig:kld}
\end{wrapfigure}

\textbf{Assumption 4.}
\emph{We choose an isotropic Gaussian weight prior} $p(w|\sigma^2)=\prod \calN(w_{ij}|0, \sigma^2)$.
This allows us to analytically compute the KL-divergence of the variational distribution and the prior:
\begin{equation}
  \KL\big[q(w|\theta)\big\|p(w)\big] = \sum_{i,j} \frac{1}{2}\left(\log\frac{\sigma^2}{\alpha_j w_{ij}^2} + \frac{(1 + \alpha_j)w_{ij}^2}{\sigma^2} - 1\right).
\end{equation}
This leads to a different implicit prior for the weight means, as can be seen in \cref{fig:kld}. 
While for a mean-field variational distribution the Kullback-Leibler divergence encourages weight means close to zero, the variational distribution induced by multiplicative activation noise favors small but non-zero weight means where the optimal size is dependent on the variance 
of the activation noise $\alpha$ and prior variance $\sigma^2$ and discourages sign changes, leading to different regularization of the network's weights.

Note that $\KL[q(w|\theta)||p(w)]$ converges to zero for $\alpha_j\rightarrow\infty$, $w_{ij}^2\rightarrow 0$, and $\alpha_j w_{ij}^2 \rightarrow \sigma^2$ for all $i,j$.
Therefore, by minimizing the KL divergence term the activation noise is increased while the influence on the weights resembles that of weight decay.
Further, for fixed $\alpha > 0$ the KL divergence is strongly convex with respect to $w^2>0$ and obtains its minimum at $\frac{\sigma^2}{1+\alpha}$, so the weight decay effect is stronger for weights where the activation noise $\alpha_j$ is high.

For learning a tempered posterior \cite{walker2001bayesian,zhang2006}, we will allow rescaling of the Kullback-Leibler divergence term, so the generalized objective becomes
\begin{equation}
  \min_\theta -\erw_{q(w|\theta)}\big[\log p(Y|X,w)\big] + \kappa \KL\big[q(w|\theta)\big\|p(w)\big]
\end{equation}
for some $\kappa > 0$.
Bayesian inference algorithms for deep networks often use $\kappa \in (0,1)$, which corresponds to a cold posterior \cite{wenzel2020good}.

\subsection{Predictive distribution}

Similar to the log-likelihood, the predictive distribution can be computed analytically given the output distribution of the network for a Gaussian predictive distribution, while for a categorical distribution it has to be approximated.
Following the calculations in \citep{wu2019deterministic} for the Gaussian predictive distribution, we obtain
\begin{equation}
  p(y|x) \approx \calN\left(\mu(x), v_{\mu(x)} + e^{c(x) + \frac{1}{2}v_c(x)}\right).
\end{equation}

For classification, the class probability vector $\mathbf{p}$ of the categorical distribution can be estimated by sampling from the output distribution:
\begin{equation}
      \mathbf{p}(y | x) \approx \frac{1}{S}\sum_{s=1}^S \softmax(h^s(x)), \quad h^s(x) \sim \tilde q(h(x)).
\end{equation}

\section{Experiments}

\subsection{Regression on the UCI datasets}
We first test our approach on the UCI regression datasets \cite{Dua:2019} in the standard setting of a fully-connected network with one hidden layer of width 50 and 10-fold cross validation. Please refer to \cref{sec: implementation details,sec: uci settings} for details on our implementation of MNVI and information on the training settings. As can be seen in \cref{tab: uci}, our MNVI method retains most of the predictive performance of the recent sampling-free DVI and dDVI \cite{wu2019deterministic}, which propagate the full covariance matrix or only variances in a Bayesian network with a mean-field Gaussian variational posterior, while ours is more lightweight.
When comparing against other VI approaches, such as Monte Carlo mean-field VI \cite{blundell2015weight, graves2011practical} and Monte Carlo dropout \cite{gal2016dropout}, as well as ensembling \cite{lakshminarayanan2017simple}, the best performing method is dataset dependent, but our proposed MNVI generally obtains competitive results.

\begin{table}[ht]
\caption {Comparison of the average log-likelihood and standard deviation for  multiple Bayesian inference methods on the UCI regression datasets. We use the MC-MFVI implementation of \cite{wu2019deterministic}.}
\label{tab: uci}
\scriptsize
\begin{tabularx}{\textwidth}{@{}X@{}c@{\hspace{1.25em}}c@{\hspace{1.25em}}c@{\hspace{1.25em}}c@{\hspace{1.25em}}c@{\hspace{1.25em}}c@{\hspace{1.25em}}c@{}}
\toprule
& boston & concrete & energy & kin8 & power & wine & yacht\\\midrule
MC-MFVI \cite{blundell2015weight, graves2011practical} & \num{-2.43 \pm 0.03} & \bfseries \num{-3.04 \pm 0.02}  & \num{-2.38 \pm 0.02}  & \bfseries \num{2.40 \pm 0.05}  & \bfseries \num{-2.66 \pm 0.01} & \bfseries  \num{-0.78 \pm 0.02} & \num{-1.68 \pm 0.04} \\
MC Dropout \cite{gal2016dropout} & \num{-2.46 \pm 0.25} & \bfseries \num{-3.04 \pm 0.09}  & \num{-1.99\pm 0.09} & \num{0.95 \pm 0.03} & \num{-2.89 \pm 0.01} & \num{-0.93 \pm 0.05} & \num{-1.55 \pm 0.12} \\
Ensemble \cite{lakshminarayanan2017simple} & \bfseries \num{-2.41 \pm 0.25} & \num{-3.06 \pm 0.18} & \num{-1.38 \pm 0.22} & \num{1.20 \pm 0.02} & \num{-2.79 \pm 0.04} & \num{-0.94 \pm 0.12} & \num{-1.18 \pm 0.21} \\
DVI \cite{wu2019deterministic} & \bfseries \num{-2.41 \pm 0.02} & \num{-3.06 \pm 0.02} & \bfseries \num{-1.01 \pm 0.06} & \num{1.13 \pm 0.00} & \num{-2.80 \pm 0.00} & \num{-0.90 \pm 0.01} & \num{-0.47 \pm 0.03} \\
dDVI \cite{wu2019deterministic} & \num{-2.42 \pm 0.02} & \num{-3.07 \pm 0.02} & \num{-1.06 \pm 0.06} & \num{1.13 \pm 0.00} & \num{-2.80 \pm 0.00} & \num{-0.91 \pm 0.02} & \num{-0.47 \pm 0.03} \\\midrule
MNVI (ours) &  \num{-2.43 \pm 0.02}  & \num{-3.05 \pm 0.01}  & \num{-1.33 \pm 0.05} & \num{1.15 \pm 0.01} & \num{-2.86 \pm 0.00} & \num{-0.96 \pm 0.01} & \bfseries \num{-0.37 \pm 0.02} \\\bottomrule 
\end{tabularx}
\vspace{-4mm}
\end{table}

\subsection{Image classification}
We train  a LeNet on MNIST \cite{lecun1998gradient}, an All Convolutional Network \cite{Springenberg15AllCNN} on  CIFAR-10 \cite{krizhevsky2009cifar}, and a ResNet-18 on CIFAR-10, CIFAR-100 \cite{krizhevsky2009cifar}, and ImageNet \cite{imagenet_cvpr09}.
See \cref{sec: image classification settings} for details on the training setup.
We compare the predictive performance of our proposed sampling-free variational inference scheme for networks with multiplicative activation noise to Monte Carlo Dropout \cite{gal2016dropout} with 8 samples computed at inference time, as well as sampling-free variational inference for a Gaussian mean field posterior (MFVI) similar to \cite{kandemir2018sampling,roth2016variational,wu2019deterministic}.
We evaluate its uncertainty estimates with respect to the expected calibration error (ECE) \cite{guo2017calibration} for 20 equidistantly spaced bins, the area under the misclassification-rejection curve (AUMRC), where uncertain examples are rejected based on the predictive entropy, and the misclassification rate at different rejection rates (see \cref{sec:evaluation-metrics} for details). 
As a further baseline, the accuracy of a standard deterministic network is evaluated. 
We also report the average required time for computing a single forward-pass with a batch-size of 100 for all methods on a NVIDIA GTX 1080 Ti GPU as well as the number of parameters.

As can be seen in \cref{tab:classification}, the proposed MNVI can be applied to various image classification problems, scaling up to ImageNet \cite{imagenet_cvpr09}.
Our approach is able to substantially improve both classification accuracy and uncertainty metrics compared to the deterministic baseline in all settings but classification on MNIST, where it matches the classification accuracy of a deterministic network.
Compared to the popular approximate Bayesian inference method MC Dropout \cite{gal2016dropout}, MNVI matches the accuracy on MNIST while significantly increasing the accuracy for ResNet18 on CIFAR-10 and -100.
While MC-Dropout produces better ECE for both architectures on CIFAR-10, MNVI achieves lower calibration error on MNIST and CIFAR-100 as well as lower AUMRC in three out of four settings.
Sampling-free variational inference with a mean field posterior approximation (MFVI), while having almost double the number of parameters and slightly higher inference time, only achieves a slightly worse misclassication rate than MNVI.
More importantly, for all settings MNVI has better calibrated predictions as indicated by the lower ECE and better separates false from correct predictions resulting in a lower AUMRC in three out of four setting.
Thus, our proposed method has significantly fewer parameters than MFVI and is faster than MC Dropout, scaling up all the way to ImageNet, while matching or even surpassing the accuracy and uncertainty estimation of both methods. 

\begin{table}[t]
\caption{Comparison for different image classification datasets (lower is better for all metrics).}
\label{tab:classification}
\tiny
\begin{tabularx}{\textwidth}{@{}l@{\hspace{0.85em}}X@{}c@{\hspace{0.85em}}c@{\hspace{0.85em}}c@{\hspace{0.85em}}c@{\hspace{0.85em}}c@{\hspace{0.85em}}c@{\hspace{0.85em}}c@{\hspace{0.85em}}c@{\hspace{0.85em}}c@{}}
\toprule 
& & Misclass. & NLLH & ECE & AUMRC  & MR10\% & MR25\% & MR50\% & Inference & Parameters \\
& & [\%] & & & & [\%] & [\%] & [\%] & [ms]\\\midrule
& Deterministic & \bf{0.55}\scalebox{.65}{$[\pm$0.06$]$} & 0.0266\scalebox{.65}{$[\pm$0.0007$]$}  & 4.03\scalebox{.65}{$[\pm$0.76$]$} $\times\text{10}^\text{--3}$  & 9.91\scalebox{.65}{$[\pm$1.23$]$} $\times\text{10}^\text{--5}$ & -- & -- & -- & 0.86\scalebox{.65}{$[\pm$0.16$]$} & \num{1.111e6}\\ 
MNIST & MC Dropout & 0.59\scalebox{.65}{$[\pm$0.01$]$}  & 0.0205\scalebox{.65}{$[\pm$0.0015$]$} & 2.64\scalebox{.65}{$[\pm$0.22$]$} $\times\text{10}^\text{--3}$ & 10.42\scalebox{.65}{$[\pm$3.03$]$} $\times\text{10}^\text{--5}$ & -- & -- & -- & 7.35\scalebox{.65}{$[\pm$0.36$]$} & \num{1.111e6}\\ 
LeNet & MFVI & 0.57\scalebox{.65}{$[\pm$0.04$]$} & \bf{0.0173}\scalebox{.65}{$[\pm$0.0008$]$} & 2.05\scalebox{.65}{$[\pm$0.37$]$} $\times\text{10}^\text{--3}$ & \bfseries 8.30\scalebox{.65}{$[\pm$0.62$]$} $\times\text{10}^\text{--5}$ & -- & -- & -- & 6.28\scalebox{.65}{$[\pm$0.45$]$} & \num{2.224e6}\\ 
& MNVI (ours) & \bf{0.55}\scalebox{.65}{$[\pm$0.03$]$} & 0.0177\scalebox{.65}{$[\pm$0.0008$]$} & \bfseries 1.91\scalebox{.65}{$[\pm$0.30$]$} $\times\text{10}^\text{--3}$ &  8.33\scalebox{.65}{$[\pm$0.64$]$} $\times\text{10}^\text{--5}$ & -- & -- & -- & 5.97\scalebox{.65}{$[\pm$0.63$]$} & \num{1.114e6}\\\midrule

& Deterministic & 7.97\scalebox{.65}{$[\pm$0.20$]$} & 0.4278\scalebox{.65}{$[\pm$0.0088$]$} & 0.0574\scalebox{.65}{$[\pm$0.0025$]$} & 0.00942\scalebox{.65}{$[\pm$0.00026$]$} & 3.54\scalebox{.65}{$[\pm$0.04$]$} & 0.87\scalebox{.65}{$[\pm$0.08$]$} & -- & 1.84\scalebox{.65}{$[\pm$0.11$]$} & \num{1.370e6}\\
CIFAR-10& MC Dropout & \bf{7.16}\scalebox{.65}{$[\pm$0.25$]$} & \bf{0.2567}\scalebox{.65}{$[\pm$0.0032$]$} & \bf{0.0276}\scalebox{.65}{$[\pm$0.0014$]$} & \bf{0.00812}\scalebox{.65}{$[\pm$0.00014$]$} & \bf{3.07}\scalebox{.65}{$[\pm$0.10$]$} & \bf{0.73}\scalebox{.65}{$[\pm$0.11$]$} & -- & 11.86\scalebox{.65}{$[\pm$0.55$]$} & \num{1.370e6}\\
AllCNN & MFVI & 7.72\scalebox{.65}{$[\pm$0.14$]$} & 0.3483\scalebox{.65}{$[\pm$0.0064$]$} & 0.0495\scalebox{.65}{$[\pm$0.0007$]$} & 0.00898\scalebox{.65}{$[\pm$0.00044$]$} & 3.40\scalebox{.65}{$[\pm$0.17$]$} & 0.81\scalebox{.65}{$[\pm$0.09$]$} & -- & 12.75\scalebox{.65}{$[\pm$0.64$]$} & \num{2.739e6}\\
& MNVI (ours) & 7.62\scalebox{.65}{$[\pm$0.35$]$} & 0.3522\scalebox{.65}{$[\pm$0.0150$]$} & 0.0492\scalebox{.65}{$[\pm$0.0033$]$} & 0.00895\scalebox{.65}{$[\pm$0.00057$]$} & 3.42\scalebox{.65}{$[\pm$0.09$]$} & 0.83\scalebox{.65}{$[\pm$0.07$]$} & -- & 9.96\scalebox{.65}{$[\pm$0.52$]$} & \num{2.739e6}\\
\midrule

& Deterministic & 5.94\scalebox{.65}{$[\pm$0.26$]$} & 0.2605\scalebox{.65}{$[\pm$0.0055$]$} & 0.0386\scalebox{.65}{$[\pm$0.0021$]$} & 0.00623\scalebox{.65}{$[\pm$0.00066$]$} & 2.05\scalebox{.65}{$[\pm$0.16$]$} & 0.53\scalebox{.65}{$[\pm$0.07$]$} & -- & 5.64\scalebox{.65}{$[\pm$0.23$]$} & \num{11.17e6}\\
CIFAR-10 & MC Dropout & 5.70\scalebox{.65}{$[\pm$0.09$]$} & \bf{0.2185}\scalebox{.65}{$[\pm$0.0063$]$} & \bf{0.0278}\scalebox{.65}{$[\pm$0.0006$]$} & 0.00576\scalebox{.65}{$[\pm$0.00021$]$} & 1.91\scalebox{.65}{$[\pm$0.05$]$} & 0.46\scalebox{.65}{$[\pm$0.03$]$} & -- & 47.34\scalebox{.65}{$[\pm$2.12$]$} & \num{11.17e6}\\
ResNet18 & MFVI & 5.63\scalebox{.65}{$[\pm$0.21$]$} & 0.2561\scalebox{.65}{$[\pm$0.0063$]$} & 0.0372\scalebox{.65}{$[\pm$0.0011$]$} & 0.00564\scalebox{.65}{$[\pm$0.00025$]$} & \bf{1.90}\scalebox{.65}{$[\pm$0.14$]$} & \bf{0.44}\scalebox{.65}{$[\pm$0.06$]$} & -- & 34.92\scalebox{.65}{$[\pm$1.18$]$} & \num{22.34e6}\\
& MNVI (ours) & \bf{5.60}\scalebox{.65}{$[\pm$0.14$]$} & 0.2462\scalebox{.65}{$[\pm$0.0069$]$} & 0.0346\scalebox{.65}{$[\pm$0.0010$]$} & \bf{0.00553}\scalebox{.65}{$[\pm$0.00004$]$} & 1.92\scalebox{.65}{$[\pm$0.09$]$} & \bf{0.44}\scalebox{.65}{$[\pm$0.07$]$} & -- & 33.82\scalebox{.65}{$[\pm$0.66$]$} & \num{11.18e6}\\
\midrule 

& Deterministic & 27.38\scalebox{.65}{$[\pm$0.57$]$} & 1.266\scalebox{.65}{$[\pm$0.019$]$} & 0.133\scalebox{.65}{$[\pm$0.004$]$} & 0.0823\scalebox{.65}{$[\pm$0.0014$]$} & 20.94\scalebox{.65}{$[\pm$0.33$]$} & 14.29\scalebox{.65}{$[\pm$0.16$]$} & 4.71\scalebox{.65}{$[\pm$0.28$]$} & 5.75\scalebox{.65}{$[\pm$0.25$]$} & \num{11.22e6}\\ 
CIFAR-100& MC Dropout & 27.87\scalebox{.65}{$[\pm$0.37$]$} & 1.240\scalebox{.65}{$[\pm$0.005$]$} &  0.116\scalebox{.65}{$[\pm$0.005$]$} & 0.0830\scalebox{.65}{$[\pm$0.0002$]$} & 22.22\scalebox{.65}{$[\pm$0.37$]$} & 14.63\scalebox{.65}{$[\pm$0.20$]$} & 4.68\scalebox{.65}{$[\pm$0.24$]$} & 46.75\scalebox{.65}{$[\pm$2.21$]$} & \num{11.22e6}\\
ResNet18& MFVI & 26.91\scalebox{.65}{$[\pm$0.10$]$} & 1.271\scalebox{.65}{$[\pm$0.016$]$} &  0.131\scalebox{.65}{$[\pm$0.003$]$} & 0.0787\scalebox{.65}{$[\pm$0.0003$]$} & 21.30\scalebox{.65}{$[\pm$0.12$]$} & 13.57\scalebox{.65}{$[\pm$0.08$]$} & 4.48\scalebox{.65}{$[\pm$0.15$]$} & 34.71\scalebox{.65}{$[\pm$0.86$]$} & \num{22.43e6}\\
& MNVI (ours) & \bf{25.30}\scalebox{.65}{$[\pm$0.50$]$} & \bf{1.085}\scalebox{.65}{$[\pm$0.011$]$} & \bf{0.105}\scalebox{.65}{$[\pm$0.005$]$} & \bf{0.0740}\scalebox{.65}{$[\pm$0.0017$]$} & \bf{20.01}\scalebox{.65}{$[\pm$0.43$]$} & \bf{12.82}\scalebox{.65}{$[\pm$0.39$]$} & \bf{3.90}\scalebox{.65}{$[\pm$0.37$]$} & 34.02\scalebox{.65}{$[\pm$0.83$]$} & \num{11.23e6}\\\midrule 
ImageNet & Deterministic &  31.09 &  1.282 & \bfseries 0.0313 &  0.1106 &  25.96 & 18.92 & 8.36 & 8.63\scalebox{.65}{$[\pm$0.59$]$} & \num{11.69e6}\\ 
ResNet18 & MNVI (ours) & \bfseries 31.05 & \bfseries 1.276 & 0.0388 & \bfseries 0.1092 & \bfseries 25.78 & \bfseries 18.68 & \bfseries 7.96 & 43.75\scalebox{.65}{$[\pm$1.06$]$} & \num{11.69e6}\\\bottomrule 
\end{tabularx}
\end{table}

\section{Conclusion}
In this paper, we proposed to use the distribution induced by multiplicative Gaussian activation noise as a posterior approximation for sampling-free variational inference in Bayesian neural networks.
The benefits of this variational posterior are a reduction of the number of parameters and required computation.
Our experiments show that the suggested posterior approximation retains or even improves over the accuracy of the Gaussian mean-field posterior approximation, while requiring a negligible amount of additional parameters compared to a deterministic network or MC dropout.
Our approach can be successfully applied to train Bayesian neural networks for various image classification tasks, matching or even surpassing the predictive accuracy of deterministic neural networks while producing better calibrated uncertainty estimates.
Because of these promising results, we hope that this as well as further research on efficient sampling-free variational inference methods will lead to a more widespread adoption of Bayesian neural networks in practice.

\section*{Acknowledgements}

This project has received funding from the European Research Council (ERC) under the European Union’s Horizon 2020 research and innovation programme (grant agreement No 866008).

\bibliographystyle{abbrvnat}
\bibliography{refs}

\newpage

\appendix

\section{Implementation Details}
\label{sec: implementation details}
For a numerical stable implementation of sampling-free variational inference, some attention to critical operations is needed.
To model the variance of the multiplicative Gaussian activation noise, we use variables $\rho_j \in \bbR$ and map them to $\alpha_j = \log(1 + \exp(\rho_j))$. 
We initialize $\rho_j$ as $-3.0$, which equals an initial variance of $\alpha_j \approx 0.02$.
For calculating the KL-divergence, we add a constant $\eps=10^{-10}$ to the term inside the logarithm.
Further, we bound the variance of the input distribution for ReLU activations from below by $10^{-5}$.
During training, we use gradient clipping if the $\infty$-norm of the gradient is greater than 0.1 for the classification experiments or 1.0 for the regression experiments.

\section{Training and Evaluation Set-up for Regression Experiments on the UCI Datasets}
\label{sec: uci settings}

We train for 200 epochs using SGD with a learning rate of 0.05 and momentum of 0.9 for all regression experiments on the UCI datasets.\footnote{\url{https://archive.ics.uci.edu/ml/datasets.php}} We set $\kappa=1$ for the final 50 epochs but rescale the KL-divergence term by the factor 0.01 for the first 100 epochs and 0.1 from epoch 100 to 150. We consider batch sizes of 64 and 128 and prior variances in $\{\num{e0}, \num{e1}, \num{e2}, \num{e3}\}$.
The hyperparameters used for the reported results are summarized in \cref{tab: uci settings}.
To train and evaluate the models, we use 10-fold cross validation with all data points. The reported results are the average of 20 independent runs.

\begin{table}[ht]
\caption {Hyperparameter settings and number of data points $N$ for the regression experiments on the UCI datasets.}
\label{tab: uci settings}
\scriptsize
\vspace{-1mm}
\begin{tabularx}{\textwidth}{@{}lYYYYYYY@{}}
\toprule
& boston & concrete & energy & kin8 & power & wine & yacht\\\midrule
Batch size  & 64 & 64 & 64 & 128 & 128 & 128 & 64 \\
Prior variance & \num{e1} & \num{e1} & \num{e1} & \num{e1} & \num{e1} & \num{e1} & \num{e2}  \\
$N$ & 506 & 1030 & 768 & 8192 & 9568 & 1588 & 308  \\\bottomrule 
\end{tabularx}
\end{table}

\section{Training and Evaluation Set-up for Image Classification Experiments}
\label{sec: image classification settings}
\textbf{Optimizer and hyperparameter settings.}
We use AMSGrad \citeA{reddi2019convergence} to train the networks on MNIST and Nesterov-SGD with a momentum of $0.9$ for all other datasets.
The hyperparameters of the optimizer are tuned on the deterministic network.
For the BNNs, only the hyperparameters specific to them are adjusted.
We apply dropout to every layer but the last of LeNet, for the AllCNN after the first two blocks consisting of three convolutions, and after every residual block for the ResNet18.
We considered dropout probabilities $\{0.5,\ 0.1,\ 0.05,\ 0.01\}$, rescaling of the Kullback-Leibler term such that $\nicefrac{\kappa}{N} \in \{\num{2e-7}, \num{1e-7}, \num{5e-8}, \num{2e-8}\}$
for MNVI on MNIST and CIFAR-10/-100, $\nicefrac{\kappa}{N} \in \{\num{2e-8}, \num{1e-9}, \num{5e-9}, \num{2e-9}\}$ for MNVI on ImageNet,  $\nicefrac{\kappa}{N} \in \{\num{2e-6}, \num{1e-6}, \num{5e-6}, \num{2e-7}\}$ for MFVI, as well as prior variances ${\sigma^2 \in \{\num{e-2}, \num{e-3}, \num{e-4}, \num{e-5}\}}$.
A more fine-grained search for the prior variance may further improve the performance of MNVI and MFVI.
The hyperparameter settings used for the reported results are summarized in \cref{tab:hyperparams}.
Note that for MNVI and MFVI, $L_2$-regularization on the weights is replaced with the KL-term, hence the weight decay is disabled.

\begin{table}[t]
\caption{Hyperparamter settings for the classification experiments. We report the learning rate (LR), weight decay (WD), batch size (B), learning rate decay $\gamma$, the epochs at which the learning rate is reduced, total number of epochs, dropout probability $p$, as well as hyperparameters specific to MNVI and MFVI. $N$ refers to the number of training images.}
\vspace{-1mm}
\label{tab:hyperparams}
\tiny
\begin{tabularx}{\textwidth}{@{}X@{}cccccccccccc@{}}
\toprule 
& LR & WD & B & $\gamma$ & Milestones & Epochs & $p$ & $\nicefrac{\kappa_\text{MN}}{N}$ & $\sigma_\text{MN}^2$ & $\nicefrac{\kappa_\text{MF}}{N}$ & $\sigma_\text{MF}^2$ & $\rho_\text{MF}$ \\ \midrule
LeNet / MNIST & 0.001 & \num{1e-5} & 64 & 0.1 & 10 & 20 & 0.1 & \num{2e-7} & \num{e-3} & \num{5e-7} & \num{e-3} & \num{-10}\\
AllCNN / CIFAR-10 & 0.05 & \num{1e-4} & 128 & 0.1 & 150, 200, 225 & 250 & 0.1 & \num{2e-8} & \num{e-4} & \num{2e-8} & \num{e-3} & \num{-10}\\
ResNet18 / CIFAR-10 & 0.1 & \num{1e-4} & 128 & 0.2 & 60, 120, 160 & 200 & 0.05 & \num{5e-8} & \num{e-4} & \num{2e-8} & \num{e-4} & \num{-20} \\
ResNet18 / CIFAR-100 & 0.1 & \num{1e-4} & 128 & 0.2 & 60, 120, 160 & 200 & 0.05 & \num{5e-8} & \num{e-4} & \num{2e-8} & \num{e-4} & \num{-20} \\
ResNet18 / ImageNet & 0.05 & \num{1e-4} & 128 & 0.2 & 20, 40 & 60 & -- & \num{2e-9} & \num{e-5} & -- & -- & -- \\\bottomrule 
\end{tabularx}
\end{table}

\textbf{Data augmentation.}
For all image classification tasks, the input images were normalized. No additional data augmentation was used for MNIST.
For CIFAR-10/-100 we apply random rotations by up to 5 degrees and translations by up to 4 pixels to the training images, as well as randomly rescaling them by up to $10\%$. Further, brightness, contrast, saturation, and gamma of the training images are randomly perturbed by up to $0.2$ except for gamma which is randomly scaled with a multiplier within the range from $0.9$ to $1.1$.
For ImageNet, we first rescale the images such that the smaller dimension measures 256 pixels and then crop $224 \times 224$ pixel images randomly during training and from the center of the image during inference. 

\textbf{Data splits for MNIST.\footnote{\url{http://yann.lecun.com/exdb/mnist/}}} We split 5\,000 images from the 60\,000 training images for validation of the hyperparameters and early stopping, and evaluate on the 10\,000 test images.

\textbf{Data splits for CIFAR-10/-100.\footnote{\url{https://www.cs.toronto.edu/~kriz/cifar.html}}} For both datasets we use the 45\,000 of the respective 50\,000 training images for training, validate on the remaining 5\,000 training images, and use the 10\,000 test images for evaluation.

\textbf{Data splits for ImageNet.\footnote{\url{http://www.image-net.org/download-images}}} We use the ILSVRC2012 \citeA{ILSVRC} training split, containing 1.3 million images for training, and evaluate on the 50\,000 ILSVRC2012 validation images.

We run and evaluate all image classification experiments four times, except the ImageNet experiments, which are run only once for computational reasons.

\section{Evaluation Metrics}
\label{sec:evaluation-metrics}
\textbf{Expected Calibration Error.} To calculate the expected calibration error \cite{guo2017calibration}, we divide the interval $[0,1]$ into $K=20$ bins of equal length and sort predictions on the evaluation set of size $N$ into bins based on their confidence. For each bin of size $B_i$, we calculate the average confidence $\text{conf}_i$ and accuracy $\text{acc}_i$ of predictions within the bin. The ECE can then be computed as
\begin{equation}
    \text{ECE} = \sum_{i=1}^K \frac{B_i}{N} |\text{conf}_i -\text{acc}_i|.
\end{equation}

\textbf{Area Under Misclassification-Rejection-Curve.} The AUMRC metric we report is defined analogous to the AU-ARC metric \citeA{nadeem2009accuracy}, but for the misclassification rate.  We sort predictions on the evaluation set by their predictive entropy in descending order.
At each step $i$, we remove the remaining data point with the highest predictive entropy from the evaluation set and add the point $(x_i,y_i)$ to the Misclassification-Rejection-Curve, where $x_i$ is the percentage of data removed from the evaluation set and $y_i$ is the misclassification rate for the remaining data points. We then report the area under the resulting step function.

\bibliographystyleA{abbrvnat}
\bibliographyA{refs}

\end{document}